\documentclass[sigconf, natbib=true, sorting=none]{acmart}

\usepackage{multirow}
\usepackage{enumitem}
\usepackage{graphicx}
\usepackage{amsmath}
\usepackage{balance} 
\usepackage{algorithm}
\usepackage{algpseudocode}
\usepackage{xcolor} 
\usepackage[most]{tcolorbox} 
\usepackage{cleveref}      
\definecolor{codegray}{rgb}{0.95,0.95,0.95} 
\definecolor{logicblue}{rgb}{0.0, 0.4, 0.7} 

\newcommand{\AlgComment}[1]{\hfill{\color{gray}\footnotesize\ttfamily \# #1}}
\graphicspath{{figures/}}

\setcopyright{acmlicensed}
\copyrightyear{2018}
\acmYear{2018}
\acmDOI{XXXXXXX.XXXXXXX}
\acmConference[Conference acronym 'XX]{Make sure to enter the correct
  conference title from your rights confirmation email}{June 03--05,
  2018}{Woodstock, NY}
\acmISBN{978-1-4503-XXXX-X/2018/06}

\begin{document}

\title{Answer First, Reason Later: Aligning Search Relevance via Mode-Balanced Reinforcement Learning}

\author{Shijie Zhang}
\authornote{Core Contributors.}
\affiliation{%
  \institution{Qwen Applications Business Group, Alibaba Group}
  \city{Beijing}
  \country{China}
}
\affiliation{%
  \institution{Peking University}
  \city{Beijing}
  \country{China}
}
\email{xuexin.zsj@alibaba-inc.com}
\email{yaya@stu.pku.edu.cn}

\author{Xiang Guo}
\authornotemark[1]
\affiliation{%
  \institution{Qwen Applications Business Group, Alibaba Group}
  \city{Beijing}
  \country{China}
}
\email{yanxun.gx@alibaba-inc.com}

\author{Rujun Guo}
\affiliation{%
  \institution{Qwen Applications Business Group, Alibaba Group}
  \city{Beijing}
  \country{China}
}
\email{rujun.grj@alibaba-inc.com}

\author{Shaoyu Liu}
\affiliation{%
  \institution{Qwen Applications Business Group, Alibaba Group}
  \city{Beijing}
  \country{China}
}
\email{liushaoyu.lsy@alibaba-inc.com}

\author{Xiaozhao Wang}
\affiliation{%
  \institution{Qwen Applications Business Group, Alibaba Group}
  \city{Beijing}
  \country{China}
}
\email{orlando@alibaba-inc.com}

\author{Guanjun Jiang}
\affiliation{%
  \institution{Qwen Applications Business Group, Alibaba Group}
  \city{Beijing}
  \country{China}
}
\email{guanj.jianggj@alibaba-inc.com}

\author{Kevin Zhang}
\authornote{Corresponding author.}
\affiliation{%
  \institution{Peking University}
  \city{Beijing}
  \country{China}
}
\email{kevinzyz@stu.pku.edu.cn}




\begin{abstract}
Building a search relevance model that achieves both low latency and high performance is a long-standing challenge in the search industry. To satisfy the millisecond-level response requirements of online systems while retaining the interpretable reasoning traces of Large Language Models (LLMs), we propose a novel \textbf{Answer-First, Reason Later (AFRL)} paradigm. This paradigm requires the model to output the definitive relevance score in the very first token, followed by a structured logical explanation. Inspired by the success of reasoning models, we adopt a "Supervised Fine-Tuning (SFT) + Reinforcement Learning (RL)" pipeline to achieve AFRL. However, directly applying existing RL training often leads to \textbf{mode collapse} in the search relevance task, where the model forgets complex long-tail rules in pursuit of high rewards. From an information theory perspective: RL inherently minimizes the \textbf{Reverse KL divergence}, which tends to seek probability peaks (mode-seeking) and is prone to "reward hacking." On the other hand, SFT minimizes the \textbf{Forward KL divergence}, forcing the model to cover the data distribution (mode-covering) and effectively anchoring expert rules. Based on this insight, we propose a \textbf{Mode-Balanced Optimization} strategy, incorporating an SFT auxiliary loss into Stepwise-GRPO training to balance these two properties. Furthermore, we construct an automated instruction evolution system and a multi-stage curriculum to ensure expert-level data quality. Extensive experiments demonstrate that our 32B teacher model achieves state-of-the-art performance. Moreover, the AFRL architecture enables efficient knowledge distillation, successfully transferring expert-level logic to a 0.6B model, thereby reconciling reasoning depth with deployment latency.
\end{abstract}


\maketitle


\section{Introduction}

\begin{figure}[h]
\centering
\includegraphics[width=0.48\textwidth]{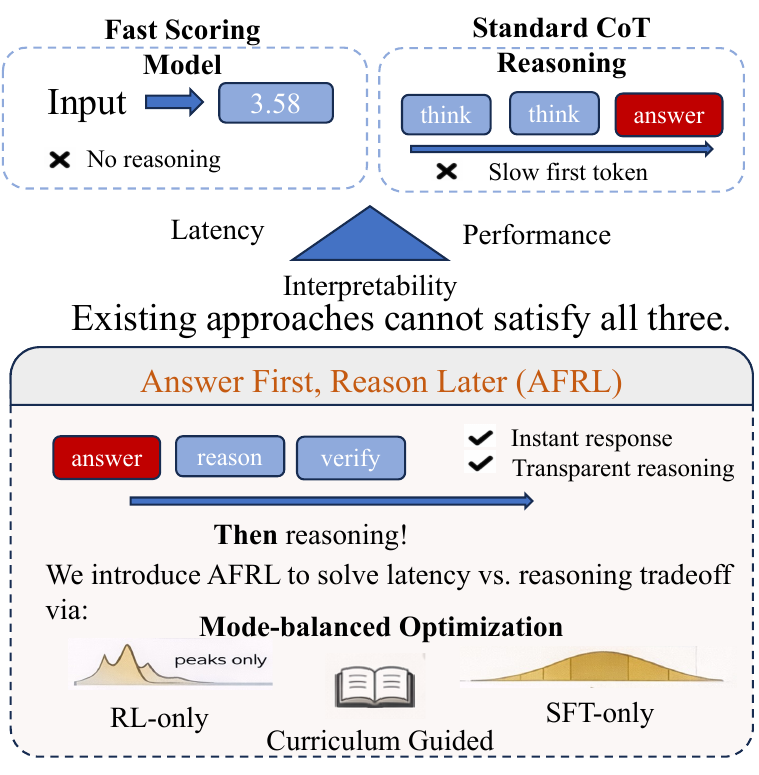}
\caption{AFRL addresses the latency–reasoning tradeoff by generating answers first and reasoning afterward, enabling instant predictions while maintaining interpretable and robust decision logic via mode-balanced optimization.}
\label{fig:teaser}
\end{figure}

Search relevance assessment serves as the fundamental engine of modern information retrieval systems, responsible for precisely mapping user intent to vast information repositories \cite{croft2010search}. In recent years, Large Language Models (LLMs) have profoundly reshaped the landscape of natural language processing through their exceptional semantic understanding and generalization capabilities \cite{achiam2023gpt, touvron2023llama}. In the field of retrieval, research focus is shifting from traditional discriminative cross-encoders \cite{devlin2019bert, liu2019roberta} to ranking paradigms based on Generative Relevance Models (GRMs). Recent works such as RankGPT~\cite{sun2023chatgpt} and RankLlama~\cite{ma2023finetuningllamamultistagetext} have demonstrated that generative LLMs can capture deep semantic alignments that are difficult for traditional models to represent.

However, deploying these models in large-scale industrial real-time systems remains a formidable challenge. Search scenarios are often constrained by the "impossible triangle" of performance, latency, and interpretability: discriminative models are responsive but operate as "black boxes" \cite{ribeiro2016should}; while the introduction of Chain-of-Thought (CoT) reasoning significantly improves predictive accuracy \cite{wei2022chain, wang2022self}, its "reason-then-answer" serial mechanism imposes a prohibitive "latency tax," the high Time-to-First-Token (TTFT) makes it impractical for online systems requiring millisecond responses \cite{pope2023efficiently, qiao2023reasoning}.

The recent emergence of reasoning models, exemplified by OpenAI o1 and DeepSeek-R1, has demonstrated remarkable "slow-thinking" capabilities through reinforcement learning (RL) \cite{guo2025deepseek, jaech2024openai}. Search relevance assessment, as a task highly dependent on the execution of expert policies, is naturally suited for this reasoning paradigm\cite{liu2025reinforcement, jin2025search, chen2025learning, wang2024reinforcement}. Inspired by this and addressing the industrial requirement for low latency, we propose the \textbf{Answer-First, Reason Later (AFRL)} paradigm. Unlike standard CoT, AFRL requires the model to output a definitive relevance score in the very first token, followed by structured logical verification. This design ensures zero-latency response for online ranking while providing a transparent window for policy debugging and RL optimization through post-hoc rationale.

A core challenge in applying Reinforcement Learning from Verifiable Rewards (RLVR) to search is optimization stability \cite{shao2024deepseekmath}. As noted in the recent work \cite{zhang2026etr}, traditional Group Relative Policy Optimization (GRPO) \cite{shao2024deepseekmath} may suffer from optimization inefficiency or even strategy collapse when handling outcome-driven learning with heterogeneous signals due to static trust region constraints. Specifically, in rule-heavy, non-mathematical tasks such as search, pure RL often triggers \textbf{mode collapse}, where models achieve high rewards through simple "shortcuts" like keyword matching while discarding complex, long-tail expert rules \cite{rafailov2023direct, geirhos2020shortcut}. From an information theory perspective, this stems from RL minimizing the Reverse KL divergence (mode-seeking), whereas Supervised Fine-Tuning (SFT) minimizes the Forward KL divergence (mode-covering) \cite{chen2025retaining}.

Addressing these issues, we propose the \textbf{Mode-Balanced Optimization} strategy, incorporating an SFT auxiliary loss as a "distributional anchor" within the Stepwise-GRPO process. Furthermore, to ensure robust learning from massive rules, we design the \textbf{PIAR (Policy Induction \& Automated Refinement)} system. Following the current self-evolving agent approaches \cite{tian2025seea, zhai2025agentevolver}, this system refines expert policy prompts through instruction self-evolution. Additionally, we enhance the efficiency gain during RL by constructing a \textbf{Multi-stage Curriculum Learning} framework that guides the model to explore from basic relevance judgments to complex long-tail scenarios\cite{wang2025reinforcement, li2024search, yu2025dynamic, zhang2025clpo}. Finally, through knowledge distillation, we successfully transfer the reasoning capabilities of our 32B teacher model to a 0.6B student model, reconciling reasoning depth with deployment efficiency.

The main contributions of this paper are:
\begin{itemize}[leftmargin=*]
    \item \textbf{Paradigm Innovation}: We propose the AFRL paradigm, decoupling reasoning capability from TTFT latency for the first time and thus satisfying both real-time industrial requirements and interpretability.
    \item \textbf{Algorithm Optimization}: Addressing mode collapse in rule-based tasks, we propose Mode-Balanced Optimization to resolve the imbalance between optimization strength and rule coverage by using both forward and reverse KL divergence.
    \item \textbf{Training Framework}: We design the PIAR system and a multi-stage curriculum learning framework, leveraging self-evolving prompts and dynamic curriculum scheduling to ensure efficient internalization of expert rules during the RL phase.
    \item \textbf{Industrial Validation}: We validate our teacher model's SOTA performance on large-scale industrial datasets and demonstrate significant gains for the 0.6B model via distillation.
\end{itemize}
\section{Preliminaries and Problem Formulation}

In this section, we formalize the search relevance assessment task as a Partially Observable Markov Decision Process (POMDP)\cite{sutton1998reinforcement, zhang2025landscape} and analyze the optimization objectives of SFT and RL.

\subsection{Relevance Assessment as a POMDP}

Industrial search relevance assessment is modeled as a sequence of decisions governed by expert policies. We formalize this process as a POMDP defined by the tuple $(\mathcal{S}, \mathcal{O}, \mathcal{A}, \mathcal{P}, R)$. The state space $\mathcal{S}$ represents the latent semantic alignment between a query and a document. The observation space $\mathcal{O}$ consists of the input text pair and explicit expert instructions. The action space $\mathcal{A}$ comprises the generated token sequence. Under our proposed AFRL paradigm, the action trajectory $\tau$ is structured as:
\begin{equation}
\tau = (y_{dec}, \mathbf{z}, y_{final})
\end{equation}
where $y_{dec}$ is the initial decision token representing an intuitive judgment of the latent state $s$, $\mathbf{z} = \{c_1, c_2, \dots, c_k\}$ is a structured logical trace containing $k$ critical checkpoints, and $y_{final}$ is the final confirmed result.

\subsection{Theoretical Analysis: Forward vs. Reverse KL}

We analyze the properties of Supervised Fine-Tuning (SFT) and Reinforcement Learning (RL) through the lens of KL divergence to justify our hybrid objective. 

SFT minimizes the \textbf{Forward KL divergence} between the expert data distribution $\pi_{data}$ and the model policy $\pi_\theta$:
\begin{equation}
\min_\theta D_{KL}(\pi_{data} \parallel \pi_\theta) = \mathbb{E}_{x,y \sim \pi_{data}} [-\log \pi_\theta(y|x)] + C
\end{equation}
The forward KL objective exhibits a \textbf{mode-covering} property (zero-forcing), where $\pi_\theta(x)$ must be non-zero whenever $\pi_{data}(x) > 0$. In our framework, SFT acts as a distributional anchor, ensuring the model covers the full support of expert rules and preventing the forgetting of long-tail logic.

Reinforcement learning minimizes the \textbf{Reverse KL divergence} between $\pi_\theta$ and an optimal distribution $\pi^* \propto \exp(R(\tau)/\eta)$\cite{chan2022greedification}:
\begin{equation}
\min_\theta D_{KL}(\pi_\theta \parallel \pi^*) = \mathbb{E}_{\tau \sim \pi_\theta} [\log \pi_\theta(\tau) - \frac{R(\tau)}{\eta}] + \log Z
\end{equation}
The reverse KL objective exhibits a \textbf{mode-seeking} property (zero-avoiding). It encourages the model to concentrate probability mass on high-reward peaks. While this sharpens decision boundaries, it risks dropping necessary but lower-probability rules (mode collapse). Our method combines these two objectives.

\subsection{Reinforcement Learning from Verifiable Rewards}

To address reward sparsity in complex reasoning tasks, we adopt Reinforcement Learning from Verifiable Rewards (RLVR)~\cite{ sutton1998reinforcement}. The reward $R(\tau)$ is decomposed into verifiable components corresponding to critical checkpoints in the logical trace:
\begin{equation}
R(\tau) = \sum_{i=1}^{k} \mathbb{I}(\text{check}(c_i, \text{rules}))
\end{equation}
This formulation allows for stepwise advantage weighting, ensuring that critical logical transitions receive the strongest optimization signals.

\section{Methodology}

\begin{figure*}[t]
\centering
\includegraphics[width=\textwidth]{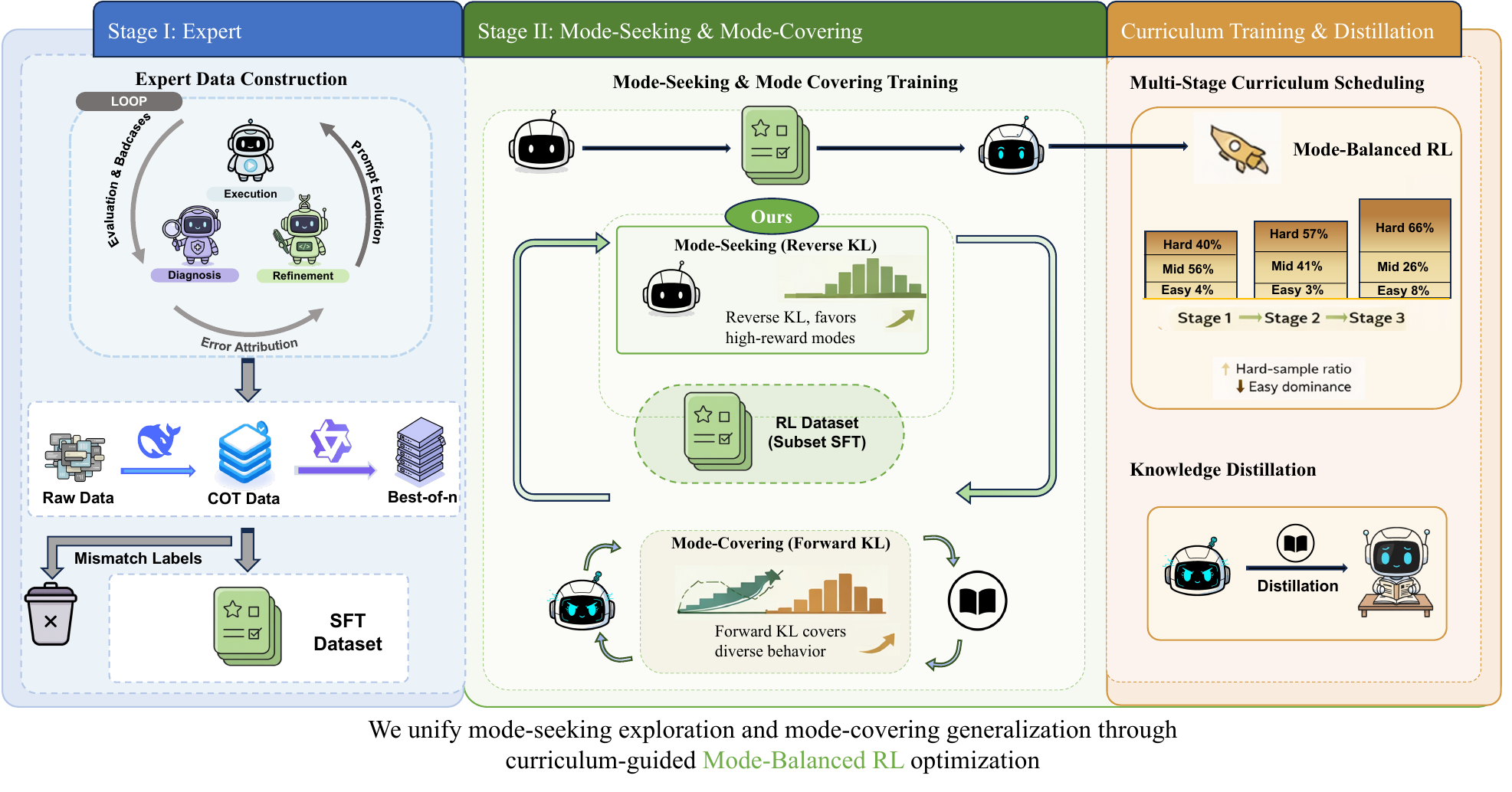}
\caption{Overview of the proposed framework. (Left) Expert Data Construction via the PIAR loop and hard-sample mining; (Middle) Mode-Balanced Optimization, a hybrid training paradigm balancing mode-seeking (Reverse KL) and mode-covering (Forward KL) dynamics; (Right) Curriculum-Guided Learning and final knowledge distillation to bridge the gap between high-level reasoning and deployment efficiency.}
\label{fig:method}
\end{figure*}

In this section, we detail the training framework designed to construct a rule-grounded expert model, as illustrated in Figure \ref{fig:method}. The framework consists of the AFRL paradigm design, automated instruction refinement, structured data synthesis, a strict reward mechanism, and a mode-balanced reinforcement learning strategy.

\subsection{AFRL: Answer-First, Reason Later Paradigm}
Based on the POMDP formulation, we re-examine the inference structure. Unlike the strict causal chains in mathematical reasoning, we posit that relevance assessment involves an immediate expert intuition followed by a logical audit. Accordingly, we propose the \textbf{Answer-First, Reason Later (AFRL)} paradigm. The action trajectory $\tau$ is defined as:
\setlength{\abovedisplayskip}{3pt}
\setlength{\belowdisplayskip}{3pt}
\begin{equation}
\tau = \{y_{dec}, \mathbf{z}_{trace}, y_{final}\}
\end{equation}
where $y_{dec}$ is the decision token generated at the very first position. This design ensures that the logits of the first token condense the global reasoning information, providing a high signal-to-noise ratio for subsequent distillation. $\mathbf{z}_{trace}$ represents the structured evidence chain containing 9 steps, where Steps 4--8 contain critical checkpoints (e.g., intent analysis, timeliness checks). Each sub-conclusion is explicitly wrapped in a \texttt{\textbackslash boxed\{\}} format for auditing. $y_{final}$ serves as the final consistency confirmation.

\begin{figure}[t]
    \centering
    \begin{tcolorbox}[
        colback=codegray, 
        colframe=black, 
        title=\textbf{AFRL Paradigm and Expert Policy Structure},
        fonttitle=\normalsize\bfseries, 
        fontupper=\small\ttfamily\linespread{1}\selectfont, 
        arc=2pt,
        outer arc=2pt,
        boxrule=0.8pt,
        left=2pt, right=2pt, top=2pt, bottom=2pt, 
        boxsep=1pt 
    ]
    \textbf{Input:} Query: "\textit{Why do plants need light?}" \quad Doc: "\textit{...Photosynthesis...}" \\
    \rule{\linewidth}{0.4pt} \\
    \textbf{Output (AFRL Sequence):} \\
    \textcolor{red}{[3]} \AlgComment{First Token Score (Zero-latency)} \\
    \textcolor{logicblue}{<think>} \\
    \textbf{Step 1: Intent} \textit{Educational/Scientific explanation.} \\
    \textbf{Step 2: Domain} \textit{Biology $\leftrightarrow$ Biology. \boxed{Yes}} \\
    \textbf{Step 3: Freshness} \textit{Static knowledge. \boxed{None}} \\
    \textbf{Step 4: Irrelevant?} \textit{Matches core entities. \boxed{No}} \\
    \textbf{Step 5: Weak?} \textit{Direct answer provided. \boxed{No}} \\
    \textbf{Step 6: Strong?} \textit{Explicitly explains photosynthesis. \boxed{Yes}} \\
    \textbf{Step 7: Premium?} \textit{Rich details on chlorophyll. \boxed{Yes}} \\
    \textbf{Step 8: Official?} \textit{Generic educational site. \boxed{No}} \\
    \textbf{Step 9: Synthesis} \textit{Content provides deep satisfying answer.} \\
    \textcolor{logicblue}{</think>} \\
    \textcolor{red}{[3]} \AlgComment{Final Confirmation}
    \end{tcolorbox}
    \label{fig:afrl_paradigm} 
\end{figure}
\subsection{Automated Instruction Refinement}

To translate unstructured business documentation into precise model instructions, we implement the Policy Induction and Automated Refinement (PIAR) system. This system operates through an Act-Diagnose-Evolve closed loop. First, the model executes inference on a validation set using the current instruction. Next, an evaluator model performs semantic error attribution on misclassified samples to identify specific misinterpreted rules. Finally, a refiner model iteratively rewrites the instructions based on these attribution reports, dynamically adjusting rule granularity and weights until performance converges.

\subsection{Structured Data Synthesis}

High-quality trajectories are the foundation of the teacher model. We construct a synthesis pipeline based on 150,000 human-annotated labels.
\begin{itemize}[leftmargin=*]
    \item \textbf{Diversity Sampling}: We utilize a strong base model Deepseek-R1\cite{guo2025deepseek} to perform diversity sampling, generating $N=10$ parallel reasoning paths for each sample.
    \item \textbf{Consistency Filtering}: We apply a strict filter to remove trajectories where the intermediate boxed conclusions or the final answer conflict with the ground truth label.
    \item \textbf{Best-of-N Selection}: To ensure logical rigor, we employ the Qwen3-Max-Thinking\cite{qwen3maxthinking} as an expert judge. It evaluates the candidate paths based on logical soundness and rule adherence, selecting the optimal trajectory to serve as the training target.
\end{itemize}

\subsection{Strict Rule-based Reward Formulation}

To enforce the \textbf{AFRL} principle, we design a deterministic reward function $R(\tau)$ that strictly penalizes correct answers derived from incorrect reasoning. Unlike standard RLHF, which uses learned reward models, we implement a rule-based verification mechanism. The total reward is formulated as a gated composition:
\begin{equation}
R(\tau) = \mathbb{I}_{fmt} \cdot \mathbb{I}_{cst} \cdot \mathbb{I}_{logic} \cdot (\alpha R_{res} + \beta R_{cot})
\end{equation}
where $\alpha=0.72$ and $\beta=0.28$. The components are defined as:

\begin{enumerate}
    \item \textbf{Strict Gating Indicators ($\mathbb{I}$)}: 
    $\mathbb{I}_{fmt}$ ensures the trajectory contains all 9 steps and correct boxed formatting. $\mathbb{I}_{cst}$ checks self-consistency between $y_{dec}$ and $y_{final}$. Crucially, $\mathbb{I}_{logic}$ is a \textbf{logic circuit breaker}. We extract the intermediate boxed results from Steps 4--8 to infer an expected answer $\hat{y}$. If $\hat{y} \neq y_{pred}$ or $\hat{y} \neq y_{gt}$, then $\mathbb{I}_{logic}=0$, zeroing out the entire reward.
    
    \item \textbf{Implicit CoT Verification ($R_{cot}$)}: 
    Instead of dense step-level labeling, we verify the CoT implicitly via the logic gate. If the gate passes ($\mathbb{I}_{logic}=1$), it implies the reasoning trace validly supports the correct conclusion, granting full $R_{cot}$.
    
    \item \textbf{Ordinal Result Reward ($R_{res}$)}: 
    Given the ordinal nature of relevance scores, we penalize deviation based on distance:
    \begin{equation}
    R_{res} = \begin{cases} 
    1.0 & \text{if } y_{pred} = y_{gt} \\
    -\gamma |y_{pred} - y_{gt}| & \text{if } y_{pred} \neq y_{gt}
    \end{cases}
    \end{equation}
\end{enumerate}

\subsection{Mode-Balanced Hybrid RL}

After a cold-start SFT phase, the model enters the core Reinforcement Learning stage. Addressing the risk where pure RL leads to "mode collapse" (forgetting complex rules to perform reward hacking), we propose a hybrid optimization framework based on \textbf{Mode-Balanced RL}.

\subsubsection{Stepwise Advantage Weighting}
We employ the Group Relative Policy Optimization (GRPO) algorithm. To address reward sparsity in long chains, we implement stepwise advantage weighting. The reward signal is mapped specifically to critical tokens ($y_{dec}$, boxed conclusions, $y_{final}$), scaling the advantage $\hat{A}_t$ at these positions to focus optimization on logical turning points.

\subsubsection{Hybrid Optimization Objective}
We formulate a total loss function that balances two information-theoretic forces:
\begin{equation}
\mathcal{L}_{Total}(\theta) = \gamma_t \mathcal{L}_{SFT}(\theta) + \alpha_t \mathcal{L}_{GRPO}(\theta)
\end{equation}

\begin{itemize}[leftmargin=*]
    \item \textbf{Mode-Seeking (Reverse KL)}: $\mathcal{L}_{GRPO}$ minimizes the reverse KL divergence. It exhibits a zero-avoiding property, driving the model to concentrate probability mass on the highest-reward paths, effectively sharpening the decision boundary.
    \item \textbf{Mode-Covering (Forward KL)}: To prevent the model from collapsing into simple heuristics, we retain the SFT loss $\mathcal{L}_{SFT}$. This minimizes forward KL divergence, exhibiting a zero-forcing property. It acts as a \textbf{distributional anchor}, compelling the model to cover the full support of expert rules.
\end{itemize}

By dynamically scheduling $\alpha_t$ and $\gamma_t$, we achieve a smooth transition from rule anchoring to logical sharpening. The complete training procedure is summarized in Algorithm \ref{alg:hybrid_grpo}.

\begin{algorithm}[h]
\caption{Hybrid Optimization via Stepwise-GRPO}
\label{alg:hybrid_grpo}
\begin{algorithmic}[1]
    \Require Policy $\pi_\theta$, Reference $\pi_{\text{ref}}$, Datasets $\mathcal{D}_{\text{RL}}, \mathcal{D}_{\text{SFT}}$, Coeffs $\alpha_t, \gamma_t$
    \Ensure Optimized Policy $\pi_\theta$

    \While{not converged}
        \State Sample batches $\mathcal{B}_{\text{RL}} \sim \mathcal{D}_{\text{RL}}$ and $\mathcal{B}_{\text{SFT}} \sim \mathcal{D}_{\text{SFT}}$
        \For{each query $x_i \in \mathcal{B}_{\text{RL}}$}
            \State Rollout trajectories $\{\tau_{i,k}\}_{k=1}^G \sim \pi_\theta(\cdot | x_i)$
            \State Compute rewards $R_{i,k}$ and group advantage $\hat{A}_{i,k}$
            \State Set $w_t \leftarrow \lambda$ \textbf{if} $t \in \{y_{\text{dec}}, \text{boxed}, y_{\text{final}}\}$ \textbf{else} $1$
        \EndFor
        \State $\mathcal{L}_{\text{total}} \leftarrow \alpha_t \mathcal{L}_{\text{RL}} + \gamma_t \mathcal{L}_{\text{SFT}}$
        \State $\theta \leftarrow \theta - \eta \nabla_\theta \mathcal{L}_{\text{total}}$
        \State Update $\alpha_t, \gamma_t$ based on curriculum stage
    \EndWhile
    \State \Return $\pi_\theta$
\end{algorithmic}
\end{algorithm}

\subsection{Multi-stage Curriculum Learning}

We design a three-stage curriculum strategy to guide the model through increasing task difficulty. 
\textbf{Stage 1 (Structure Establishment)} focuses on stabilizing the AFRL format using medium-difficulty samples. 
\textbf{Stage 2 (Logic Consolidation)} significantly increases the proportion of hard samples, leveraging GRPO to explore correct paths in complex scenarios. 
\textbf{Stage 3 (Boundary Breakthrough)} targets hard and previously failed samples with increased training epochs. This forces the policy to align precisely with expert logic at the most ambiguous decision boundaries.
\section{Experiments}

\subsection{Benchmarks and Training Strategy}
We evaluate our framework on two core benchmarks, INDUSTRY and LONGTAIL, whose label distributions are detailed in Table \ref{tab:dataset_dist}. The \textbf{INDUSTRY} dataset comprises 14,625 samples sampled from real-world search logs across various vertical domains, such as finance and gaming, assessing rule compliance in standardized head traffic. The \textbf{LONGTAIL} dataset contains 3,881 samples curated from cold-start queries with \textbf{monthly page views (PV) fewer than 3}, serving as a stress test for pure semantic reasoning in ambiguous contexts. All samples are annotated by experts using a 5-level relevance scale (0--4).

\begin{table}[htbp]
\centering
\caption{Relevance label distribution of the INDUSTRY and LONGTAIL test sets.}
\label{tab:dataset_dist}
\begin{tabular}{crrrr}
\toprule
\textbf{Relevance} & \multicolumn{2}{c}{\textbf{INDUSTRY}} & \multicolumn{2}{c}{\textbf{LONGTAIL}} \\
\cmidrule(lr){2-3} \cmidrule(lr){4-5}
\textbf{Score} & \textbf{Count} & \textbf{Ratio (\%)} & \textbf{Count} & \textbf{Ratio (\%)} \\
\midrule
0 & 3,085 & 21.09 & 964 & 24.84 \\
1 & 4,771 & 32.62 & 1,097 & 28.27 \\
2 & 5,668 & 38.76 & 1,765 & 45.48 \\
3 & 1,004 & 6.86  & 54 & 1.39 \\
4 & 97    & 0.67  & 1 & 0.02 \\
\midrule
\textbf{Total} & \textbf{14,625} & \textbf{100.00} & \textbf{3,881} & \textbf{100.00} \\
\bottomrule
\end{tabular}
\end{table}

For model development, we first conduct Supervised Fine-Tuning (SFT) on an expert-level dataset of 150,000 samples, each following the AFRL paradigm with a definitive judgment followed by a 9-step structured reasoning trace. Transitioning to the Reinforcement Learning (RL) phase, we utilize the SFT-trained model to estimate the difficulty of these 150,000 prompts. By performing 8 independent samplings per prompt, we categorize them based on their accuracy (Acc@8): \textbf{Easy} (correct counts $\ge 7$), \textbf{Medium} (2--6), and \textbf{Hard} (1). Samples with zero correct outputs are excluded to maintain reward signal reliability. As shown in Table \ref{tab:curriculum_dist}, we implement a three-stage curriculum strategy. Stage 1 focuses on structural stabilization using predominantly Medium samples. Stage 2 introduces a higher volume of Hard samples to encourage logic exploration. Stage 3 maximizes the proportion of Hard samples to sharpen the model's discriminative precision at ambiguous decision boundaries.

\begin{table}[htbp]
\centering
\caption{Data composition and difficulty distribution across the three RL stages.}
\label{tab:curriculum_dist}
\begin{tabular}{lcccc}
\toprule
\textbf{Stage} & \textbf{Easy} & \textbf{Medium} & \textbf{Hard} & \textbf{Total} \\
\midrule
Stage 1 & 2k (4.4\%) & 25k (55.6\%) & 18k (40.0\%) & 45k \\
Stage 2 & 2k (3.3\%) & 25k (41.0\%) & 34k (55.7\%) & 61k \\
Stage 3 & 5k (8.2\%) & 16k (26.2\%) & 40k (65.6\%) & 61k \\
\bottomrule
\end{tabular}
\end{table}

\subsection{Evaluation and Implementation Details}
We employ five metrics to assess model performance from multiple dimensions. \textbf{5-Class Accuracy (5-ACC)} measures the exact match rate between predicted and ground-truth labels. \textbf{2-Class Accuracy (2-ACC)} reflects practical requirements by grouping labels $\{0, 1\}$ as "Irrelevant" and $\{2, 3, 4\}$ as "Relevant." \textbf{Macro F1} evaluates balanced recognition across imbalanced classes. For ranking performance, we introduce \textbf{Pair-wise Accuracy (Pair-ACC)} to evaluate document pairs under the same query. We also report \textbf{NDCG@3} to measure top-tier sorting quality. For ranking metrics, we calculate a \textbf{Weighted Expected Score} to obtain fine-grained signals:
\begin{equation}
s = \sum_{k=0}^{4} k \cdot P(\text{token}_k | x)
\end{equation}
where $P(\text{token}_k | x)$ is the normalized probability of the $k$-th relevance token at the first output position.

The training process is executed in two primary stages. In the SFT stage, we conduct full-parameter fine-tuning using the \textbf{LLaMA-Factory} framework to inject domain-specific expert knowledge. In the RL stage, we employ the \textbf{VERL} framework for large-scale optimization, leveraging its efficient hybrid engine to maximize training throughput. All experiments were conducted on NVIDIA H20 GPU clusters. Detailed hyperparameter configurations, including learning rates, batch sizes, and KL coefficients, are provided in Appendix A.

\subsection{Main Results}

\subsubsection{Overall Performance Comparison}
We present the comprehensive evaluation results in Table \ref{tab:main_results}. Our proposed reasoning-enhanced models (AFRL paradigm) are compared against the Standard Scoring baseline, Qwen2.5-72B-Instruct. This baseline model was trained via a robust two-stage pipeline: initial training on \textbf{15 million} general relevance samples, followed by fine-tuning on the same \textbf{150,000} expert-aligned samples used in our SFT phase.

\begin{table*}[t]
\centering
\caption{Overall performance on INDUSTRY and LONGTAIL benchmarks (\%). \textbf{Standard Scoring} refers to the traditional 1-token output pipeline. \textbf{Mode-Balanced RL} denotes our proposed dual-constrained strategy. Best results among 32B models are bolded.}
\label{tab:main_results}
\renewcommand{\arraystretch}{1.2}
\setlength{\tabcolsep}{2.5pt}
\resizebox{\textwidth}{!}{
\begin{tabular}{l cccccc | cccccc}
\toprule
\multirow{2}{*}{\textbf{Model}} & \multicolumn{6}{c}{\textbf{INDUSTRY}} & \multicolumn{6}{c}{\textbf{LONGTAIL}} \\
\cmidrule(lr){2-7} \cmidrule(lr){8-13}
 & \textbf{5-ACC} & \textbf{2-ACC} & \textbf{Pair-ACC} & \textbf{NDCG@3} & \textbf{Mac-F1} & \textbf{W-F1} & \textbf{5-ACC} & \textbf{2-ACC} & \textbf{Pair-ACC} & \textbf{NDCG@3} & \textbf{Mac-F1} & \textbf{W-F1} \\
\midrule
Qwen2.5-72B-Instruct (Standard) & 68.53 & 82.07 & 82.32 & 82.60 & 65.93 & 68.36 & 65.36 & 79.51 & 72.40 & 82.73 & 67.67 & 65.35 \\
\midrule
Qwen3-32B-SFT (AFRL) & 70.34 & 82.47 & 79.43 & 80.87 & 67.06 & 69.31 & 65.03 & 78.64 & 72.06 & 81.59 & 63.34 & 63.71 \\
Qwen3-32B-GRPO & 72.75 & 84.90 & 82.11 & 82.06 & \textbf{70.08} & 72.73 & 65.10 & 78.40 & 71.49 & 82.11 & \textbf{68.91} & 65.38 \\
Qwen3-32B-Stepwise-GRPO & 73.06 & 85.02 & 81.89 & 82.00 & 68.62 & 72.71 & \textbf{66.66} & \textbf{80.19} & 71.83 & 81.82 & 67.37 & \textbf{66.35} \\
\textbf{Qwen3-32B-Mode-Balanced-RL} & \textbf{73.07} & \textbf{85.57} & \textbf{82.40} & \textbf{82.20} & 69.40 & \textbf{72.95} & 66.49 & 80.13 & \textbf{73.43} & \textbf{82.62} & 67.23 & 65.84 \\
\bottomrule
\end{tabular}
}
\end{table*}

The transition to the Answer-First, Reason Later (AFRL) paradigm fundamentally alters the model's capacity to handle semantic complexity. As shown in Table \ref{tab:main_results}, our 32B reasoning models achieve competitive or superior performance compared to the significantly larger Qwen2.5-72B-Instruct. This is particularly evident in ranking metrics on the challenging LONGTAIL dataset, where our Mode-Balanced RL achieves a Pair-ACC of 73.43\% versus the 72B model's 72.40\%, despite the latter having seen 100x more pre-training data.

During the reinforcement learning phase, we observed a critical interplay between model confidence and ranking sensitivity. Standard reinforcement learning objectives are inherently mode-seeking, often degrading ranking quality (Pair-ACC) by eliminating the probability "softness" required to distinguish subtle relevance differences. Note that the numerical difference in Pair-ACC between training logs and Table \ref{tab:main_results} stems from the inference backend: training uses vLLM for throughput, while evaluation employs standard attention for precision.

Our Mode-Balanced RL strategy effectively reconciles this tension. By anchoring the policy to the expert-level logic distribution, it prevents overconfidence collapse. Consequently, \textbf{Qwen3-32B-Mode-Balanced-RL} consistently achieves the highest ranking metrics across both benchmarks.

\subsubsection{Standard Baseline Analysis}
To quantify the benefit of our reasoning paradigm, we compare the **Standard Scoring** version of the Qwen3-32B base model against the 72B instructor in Table \ref{tab:disc_comparison}. While the 32B standard model performs respectably, it lacks the reasoning depth to handle ambiguous queries. Our AFRL strategy (Table \ref{tab:main_results}) effectively boosts the 32B model's capabilities, allowing it to surpass the 72B baseline's Pair-ACC (82.40\% vs 82.32\%) and Weighted F1 (72.95\% vs 68.36\%).

\begin{table}[htbp]
\centering
\caption{Performance comparison of Standard Scoring baselines (single-token output) on the INDUSTRY dataset (\%).}
\label{tab:disc_comparison}
\resizebox{\columnwidth}{!}{
\begin{tabular}{lcccccc}
\toprule
\textbf{Model} & \textbf{5-ACC} & \textbf{2-ACC} & \textbf{Pair-ACC} & \textbf{NDCG@3} & \textbf{Mac-F1} & \textbf{W-F1} \\
\midrule
Qwen3-32B (Standard) & 67.52 & 81.23 & 81.34 & 81.70 & 64.92 & 66.87 \\
Qwen2.5-72B-Instruct (Standard) & 68.53 & 82.07 & 82.32 & 82.60 & 65.93 & 68.36 \\
\bottomrule
\end{tabular}
}
\end{table}

\subsubsection{Class-wise Performance Analysis}
To understand the source of improvements, we analyze the per-class performance on the INDUSTRY dataset in Table \ref{tab:per_class_metrics}. Standard GRPO tends to improve macro metrics by overfitting to specific patterns, often at the cost of precision in ambiguous classes. In contrast, Mode-Balanced RL maintains a more uniform performance distribution. Notably, for the critical "Relevant" classes (Score 2 and 3), our method achieves a superior balance, with Score 2 F1 reaching 79.82\% and Score 3 Precision improving significantly to 70.63\%, confirming that the auxiliary logic-anchoring loss helps the model sharpen its precision without collapsing into trivial solutions.

\begin{table}[htbp]
\centering
\caption{Per-class breakdown of Precision (P), Recall (R), and F1-score on the INDUSTRY dataset. Scores are formatted as P / R / F1 (\%).}
\label{tab:per_class_metrics}
\resizebox{\columnwidth}{!}{
\begin{tabular}{c ccc}
\toprule
\textbf{Label} & \textbf{Qwen3-32B-SFT} & \textbf{Qwen3-32B-GRPO} & \textbf{Mode-Balanced RL} \\
\midrule
\textbf{0} & 68.67 / 79.42 / 73.65 & 69.06 / 80.16 / 74.20 & 73.14 / 67.71 / 70.32 \\
\textbf{1} & 78.56 / 46.07 / 58.08 & 68.76 / 64.54 / 66.58 & 67.53 / 67.95 / 67.74 \\
\textbf{2} & 71.54 / 84.51 / 77.48 & 80.75 / 74.88 / 77.70 & 77.84 / 81.92 / 79.82 \\
\textbf{3} & 53.83 / 78.49 / 63.86 & 63.75 / 78.98 / 70.55 & 70.63 / 65.64 / 68.04 \\
\textbf{4} & 61.62 / 62.89 / 62.24 & 75.76 / 51.55 / 61.35 & 72.86 / 52.58 / 61.08 \\
\bottomrule
\end{tabular}
}
\end{table}

\subsection{Ablation Study}

\subsubsection{Impact of Mode-Balanced Optimization}
\begin{figure*}[t]
\centering
\includegraphics[width=0.95\textwidth]{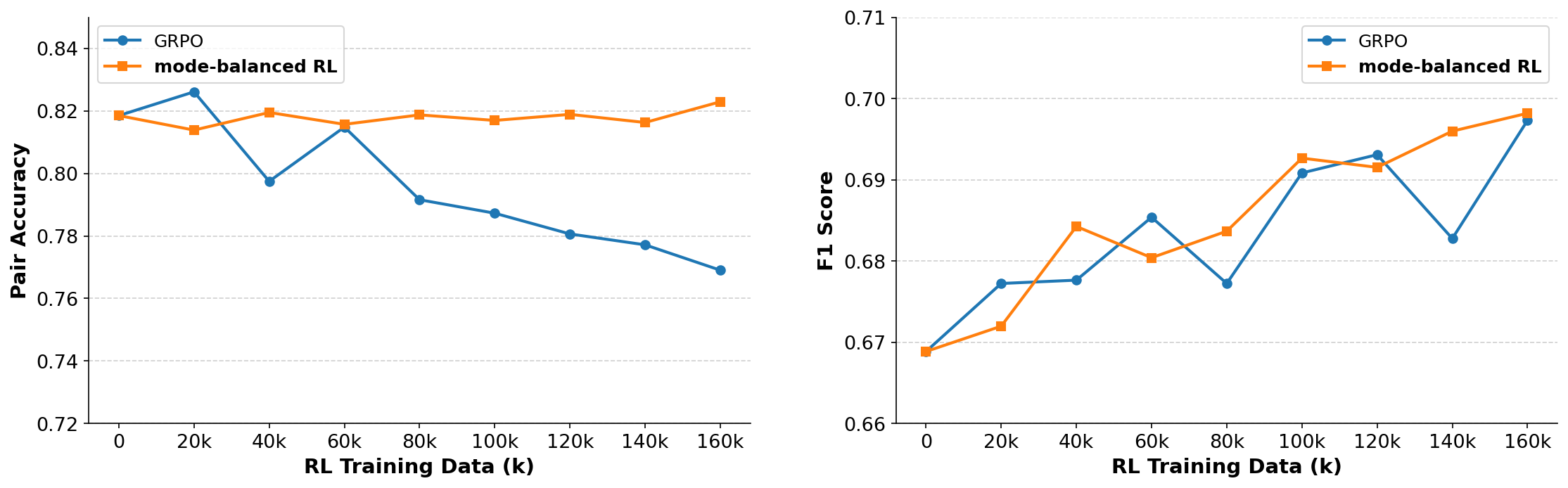}
\caption{Evolution of Pair Accuracy (left) and F1 Score (right).}
\label{fig:ablation_acc}
\end{figure*}

\begin{figure*}[t]
\centering
\includegraphics[width=0.95\textwidth]{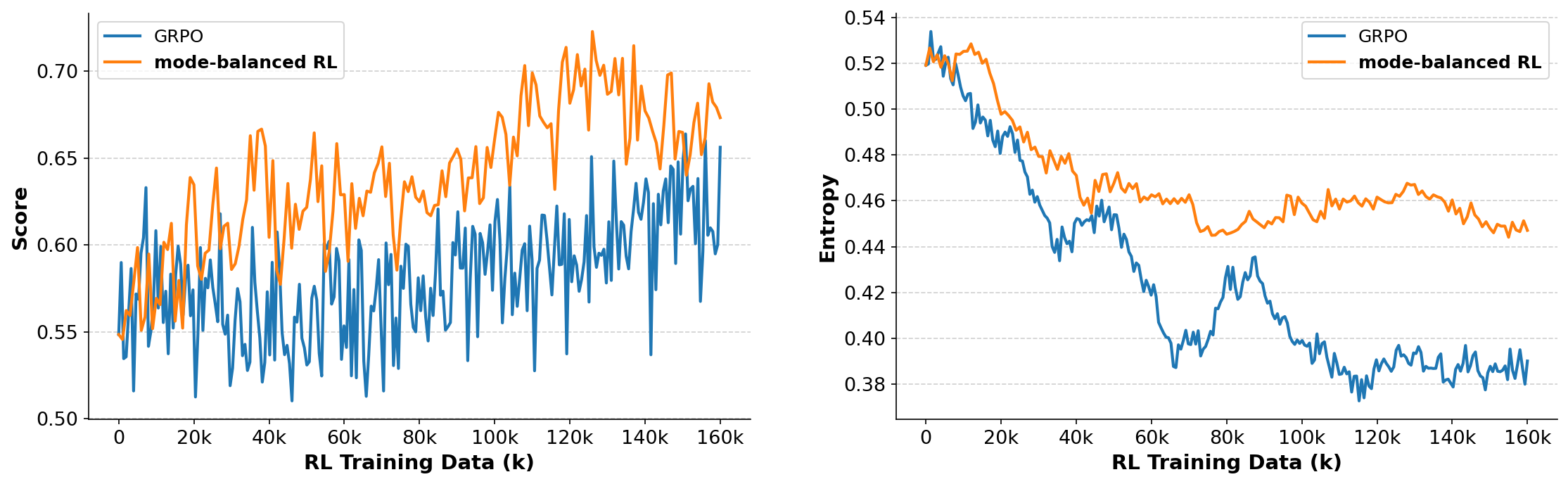}
\caption{Dynamics of Reward Score (left) and Policy Entropy (right).}
\label{fig:ablation_entropy}
\end{figure*}

We conduct a systematic ablation study to validate the effectiveness of the Mode-Balanced RL strategy. We separate the analysis into two key dimensions: ranking/classification performance (Figure \ref{fig:ablation_acc}) and training stability dynamics (Figure \ref{fig:ablation_entropy}).

\textbf{The Over-Confidence Trap in Ranking.} 
Figure \ref{fig:ablation_acc} illustrates a critical divergence between classification and ranking metrics. As training progresses, the standard GRPO (blue line) steadily improves its F1 Score, indicating it is learning to predict the correct labels. However, its Pair Accuracy suffers a catastrophic decline after the 20k mark. This phenomenon confirms the "over-confidence trap": standard RL drives the policy to maximize the probability of the winning label, causing the output distribution to approach a one-hot vector. While this benefits point-wise accuracy (F1), it eliminates the "soft" probability differences required to rank document pairs effectively. In contrast, our Mode-Balanced RL (orange line) maintains a robust upward trajectory in both metrics. By anchoring the policy to the SFT distribution, it preserves the necessary probabilistic nuance, achieving superior Pair Accuracy without sacrificing classification precision.

\textbf{Entropy and Stability Mechanism.} 
The underlying mechanism is further elucidated in Figure \ref{fig:ablation_entropy}. The Entropy plot (right) shows that standard GRPO suffers from a rapid collapse in entropy, plummeting below 0.40. This signifies that the model has collapsed into a narrow mode, losing its exploratory capability and sensitivity to ambiguous cases. Conversely, Mode-Balanced RL maintains a healthy entropy level (approx. 0.45--0.46), effectively balancing exploration (Mode-Covering) and exploitation (Mode-Seeking). This stability directly translates to the Reward Score (left), where our method exhibits a smoother, lower-variance growth curve compared to the erratic fluctuations of standard GRPO. This confirms that the auxiliary SFT loss acts as a regularizer, preventing the model from hacking the reward function via degenerate policies and ensuring consistent long-term optimization.

\subsubsection{Effectiveness of Multi-stage Curriculum Learning}
\begin{figure}[htbp]
\centering
\includegraphics[width=0.95\columnwidth]{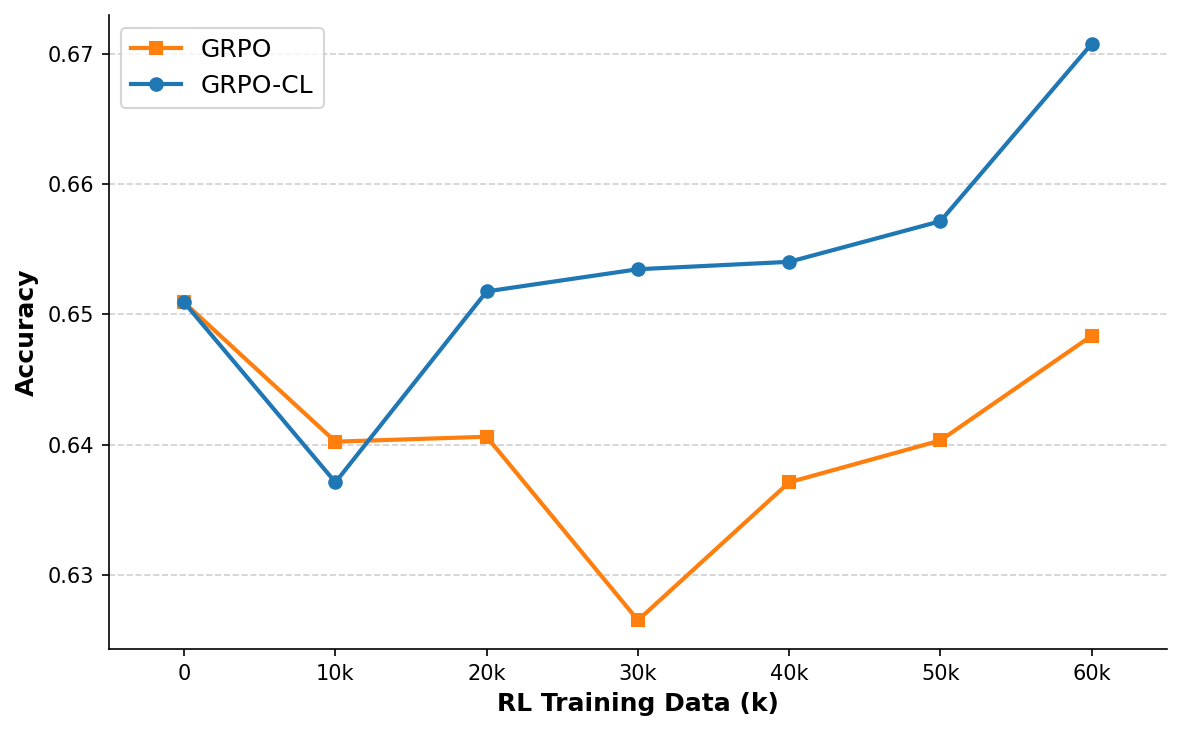} 
\caption{Comparison of accuracy evolution between standard GRPO (random sampling) and GRPO-CL (curriculum learning) on Qwen3-8B. Both methods use the same training data, differing only in the order of sample presentation.}
\label{fig:cl_ablation}
\end{figure}

To evaluate the specific impact of our curriculum learning (CL) strategy, we conducted an ablation study on the Qwen3-8B model backbone. In this experiment, we compared two training configurations using the identical RL dataset: one following our proposed multi-stage curriculum (GRPO-CL), which transitions from easy to hard samples, and the other utilizing a standard random sampling approach (GRPO). As shown in Figure \ref{fig:cl_ablation}, the curriculum learning strategy proves essential for stabilizing the reinforcement learning process in our task.

The experimental results illustrate a clear divergence in training dynamics. While both methods start from the same SFT baseline, the standard GRPO approach (orange line) exhibits significant performance fluctuations and an overall lower accuracy trajectory. This instability suggests that introducing high-difficulty or ambiguous samples prematurely can introduce excessive noise into the reward signal, leading to suboptimal policy exploration. In contrast, although the GRPO-CL (blue line) undergoes a brief adaptation period, it demonstrates a consistent and robust upward trend as training progresses. By the end of the 60k training samples, the curriculum-based approach achieves a substantial accuracy margin over the random baseline. This validates our hypothesis that a "basic-to-expert" knowledge transition allows the model to first solidify its structural reasoning before tackling complex boundary cases. Given the proven efficiency of this strategy on the 8B scale, we successfully applied the same curriculum framework to our primary 32B teacher model to achieve peak performance with optimized computational costs.

\subsection{Computational Efficiency Analysis}
We further analyze the training cost of our Mode-Balanced strategy. Intuitively, adding an auxiliary SFT loss to the RL loop introduces a small extra cost per step. However, this cost is effectively offset by the significant improvement in convergence speed.

The distribution anchoring effect provided by the SFT loss prevents the policy from getting stuck in "reward hacking" loops or suffering from early mode collapse. Consequently, each optimization step yields more robust policy improvements. Empirically, based on our reward evolution curves, the Mode-Balanced RL requires approximately 70\% of the training samples to reach the peak performance level of standard GRPO. This drastic reduction in total training iterations outweighs the slight per-step overhead, making our framework highly efficient for large-scale industrial deployment.

\subsection{Distillation Analysis}
To assess the deployment feasibility of our framework, we investigate the performance transfer from Teacher models to lightweight Student models. As shown in Table \ref{tab:distillation}, we observe distinct behaviors between the standard scoring and reasoning paradigms.

For the \textbf{Base (Standard Scoring)} configuration, the Student model achieves near-lossless replication of the Teacher's performance (68.5\% vs. 68.5\%). This suggests that the single-token probability distribution is relatively simple to fit. In contrast, under the \textbf{CoT (AFRL)} paradigm, while the Student model (70.9\%) significantly outperforms the Base Student, it exhibits a performance gap compared to its Teacher (73.1\%). This indicates that reasoning traces provide richer supervision signals boosting student capabilities beyond standard baselines.

\begin{table}[htbp]
\centering
\caption{Distillation performance comparison on the INDUSTRY dataset (5-ACC). The Base Teacher corresponds to Qwen2.5-72B (Standard), and the CoT Teacher corresponds to our Qwen3-32B (Model-Balanced-RL).}
\label{tab:distillation}
\begin{tabular}{lcc}
\toprule
\textbf{Paradigm} & \textbf{Teacher Acc} & \textbf{Student Acc} \\
\midrule
Base (Standard Scoring) & 68.5\% & 68.5\% \\
CoT (AFRL Paradigm)     & \textbf{73.1\%} & \textbf{70.9\%} \\
\bottomrule
\end{tabular}
\end{table}
\section{Related Works}

\textbf{Evolution of Search Relevance and Reasoning Paradigms.} The landscape of search relevance has transitioned from traditional discriminative cross-encoders \cite{devlin2019bert} to generative large language models (LLMs) that treat ranking as a sequence generation task \cite{sun2023chatgpt, ma2024fine, li2024survey, yan2024consolidating, pan2025llm}. While Chain-of-Thought (CoT) prompting \cite{wei2022chain} significantly enhances reasoning depth, its "reason-then-answer" structure imposes a prohibitive "latency tax," specifically high Time-to-First-Token (TTFT), rendering it impractical for real-time industrial search systems. Furthermore, traditional rankers often operate as uninterpretable black boxes, complicating policy debugging and alignment. Recent studies have explored post-hoc rationalization and "answer-first" structures \cite{chen2025distilling} to decouple final decisions from logical justifications. Our proposed AFRL paradigm extends this line of work by forcing the model to emit a definitive relevance label in the very first token, thereby satisfying millisecond-level production constraints while providing a structured logical trace that ensures high interpretability for expert policy alignment.

\textbf{Stability in Reinforcement Learning and Knowledge Transfer.} With the success of DeepSeek-R1 \cite{guo2025deepseek}, Reinforcement Learning from Verifiable Rewards (RLVR) and Group Relative Policy Optimization (GRPO)\cite{shao2024deepseekmath} have become standard paradigms for eliciting reasoning capabilities\cite{zhuang2025rank, yue2025does, plaat2025multi}. However, applying RLVR to non-mathematical tasks like search relevance often triggers "mode collapse" or "reward hacking," where models discard complex expert rules in favor of superficial shortcuts. Recent theoretical work \cite{chen2025retaining} elucidates the fundamental difference in distributional dynamics between Supervised Fine-Tuning (SFT) and Reinforcement Learning (RL): SFT minimizes Forward KL divergence, exhibiting a \textbf{mode-covering} property that maintains the breadth of the expert rule distribution; conversely, RL minimizes Reverse KL divergence, exhibiting a \textbf{mode-seeking} property that tends to converge on high-reward local peaks. To mitigate this, our Mode-Balanced Optimization incorporates an SFT auxiliary loss as a "distributional anchor" within the RLVR framework, leveraging the mode-covering nature of Forward KL to prevent the forgetting of long-tail expert policies. This optimized Qwen3-32B\cite{yang2025qwen3} teacher model not only achieves superior logical rigor but also facilitates the robust transfer of "expert intuition" to a Qwen3-0.6B\cite{yang2025qwen3} student model via knowledge distillation, reconciling high-level reasoning with extreme deployment efficiency.
\section{Conclusion}

In this work, we presented the \textbf{Answer-First, Reason Later (AFRL)} framework, effectively reconciling the tension between millisecond-level latency requirements and the deep reasoning capabilities of LLMs in industrial search systems. By decoupling the decision token from the rationale generation, AFRL achieves the low TTFT of discriminative models while retaining the interpretability and logic of generative reasoners.

Theoretically, we identified the root cause of "mode collapse" in rule-heavy tasks as the intrinsic \textbf{mode-seeking} nature of Reinforcement Learning (minimizing Reverse KL divergence). We demonstrated that pure RL tends to exploit rewards via shortcuts, neglecting complex long-tail rules. To address this, we proposed the \textbf{Mode-Balanced Optimization} strategy, which integrates an SFT auxiliary loss as a \textbf{mode-covering} distributional anchor (minimizing Forward KL divergence). This hybrid objective ensures the model sharpens its decision boundaries without sacrificing the breadth of expert rule coverage.

Supported by the PIAR automated instruction system and a Multi-stage Curriculum Learning strategy, our approach has proven highly robust. Extensive experiments show that our 32B teacher model not only outperforms the significantly larger Qwen2.5-72B baseline but also successfully transfers its "expert intuition" to a lightweight 0.6B model via distillation. This establishes a scalable paradigm for deploying reasoning-enhanced LLMs in latency-critical applications. Future work will extend this framework to multi-modal retrieval and explore dynamic compute allocation for varying query complexities.

\bibliographystyle{ACM-Reference-Format}
\bibliography{ref} 

\newpage

\appendix

\section{Theoretical Analysis}
\label{sec:appendix_theory}

\subsection{SFT: Minimizing Forward KL Divergence}
Supervised Fine-Tuning aims to maximize the likelihood of the expert policy $\pi_{\text{data}}$ under the model parameters $\theta$. Let the expert distribution be $P(x)$ and the model distribution be $Q_\theta(x)$. The maximum likelihood estimation (MLE) is equivalent to minimizing the \textbf{Forward KL Divergence}:
\begin{align}
\mathcal{L}_{\text{SFT}}(\theta) &= -\mathbb{E}_{x \sim P} [\log Q_\theta(x)] \\
&= \sum_{x} -P(x) \log Q_\theta(x)
\end{align}
We relate this to the KL divergence definition $D_{KL}(P \parallel Q) = \sum P(x) \log \frac{P(x)}{Q_\theta(x)}$:
\begin{align}
D_{KL}(P \parallel Q_\theta) &= \sum_{x} P(x) \log P(x) - \sum_{x} P(x) \log Q_\theta(x) \nonumber \\
&= -H(P) + \mathcal{L}_{\text{SFT}}(\theta)
\end{align}
Since the entropy of the expert data $H(P)$ is a constant with respect to $\theta$, minimizing the SFT loss is mathematically strictly equivalent to minimizing $D_{KL}(P \parallel Q_\theta)$.

\textbf{Property (Zero-Forcing / Mode-Covering):} 
Consider the divergence term $P(x) \log \frac{P(x)}{Q_\theta(x)}$. If $P(x) > 0$ (the expert considers $x$ a valid rule) but $Q_\theta(x) \to 0$ (the model ignores it), then $\log \frac{P(x)}{Q_\theta(x)} \to \infty$, causing the loss to explode. 
Consequently, the model is forced to assign non-zero probability to \textit{all} supports of the expert distribution. This \textbf{Mode-Covering} behavior acts as a "Distributional Anchor," preventing the catastrophic forgetting of long-tail rules.

\subsection{RL: Minimizing Reverse KL Divergence}
In Reinforcement Learning, we aim to find a policy $\pi_\theta$ that maximizes the expected reward $J(\theta) = \mathbb{E}_{x \sim \pi_\theta}[R(x)]$. In the context of entropy-regularized RL, the optimal policy $\pi^*$ is known to take the form of a Gibbs distribution:
\begin{equation}
\pi^*(x) = \frac{1}{Z} \exp\left(\frac{R(x)}{\eta}\right)
\end{equation}
where $Z$ is the partition function and $\eta$ is the temperature. RL optimization approximates this optimal policy by minimizing the \textbf{Reverse KL Divergence}:
\begin{align}
D_{KL}(\pi_\theta \parallel \pi^*) &= \sum_{x} \pi_\theta(x) \log \frac{\pi_\theta(x)}{\pi^*(x)} \nonumber \\
&= \sum_{x} \pi_\theta(x) \left[ \log \pi_\theta(x) - \log \left( \frac{e^{R(x)/\eta}}{Z} \right) \right] \nonumber \\
&= \sum_{x} \pi_\theta(x) \log \pi_\theta(x) - \frac{1}{\eta} \sum_{x} \pi_\theta(x) R(x) + \log Z \nonumber \\
&= -H(\pi_\theta) - \frac{1}{\eta} \mathbb{E}_{x \sim \pi_\theta}[R(x)] + \text{Const}
\end{align}
Thus, minimizing Reverse KL is equivalent to maximizing the reward plus an entropy bonus.

\textbf{Property (Zero-Avoiding / Mode-Seeking):} 
Consider the term $\pi_\theta(x) \log \frac{\pi_\theta(x)}{\pi^*(x)}$. If $\pi^*(x)$ is small (low reward), $\pi_\theta(x)$ is pressured to be small to avoid penalty. However, if $\pi^*(x)$ is large (high reward) at a specific mode $x_1$ and $\pi^*(x)$ is also non-zero at another mode $x_2$, the model can effectively minimize the divergence by collapsing all probability mass onto $x_1$ (where $R(x)$ is highest) while ignoring $x_2$, as $\pi_\theta(x_2)=0$ contributes zero to the sum.
This \textbf{Mode-Seeking} behavior sharpens decisions but leads to "Mode Collapse," where the model abandons complex, lower-frequency correct paths in favor of the single highest-reward shortcut.

\subsection{Mode-Balanced Optimization Formulation}
To reconcile these opposing dynamics, our total objective combines both divergences with dynamic curriculum coefficients $\alpha_t$ and $\gamma_t$:
\begin{equation}
\mathcal{L}_{\text{Total}} = \underbrace{\alpha_t D_{KL}(\pi_\theta \parallel \pi^*)}_{\text{Seek High Reward}} + \underbrace{\gamma_t D_{KL}(\pi_{\text{data}} \parallel \pi_\theta)}_{\text{Anchor to Rules}}
\end{equation}
This formulation ensures the model sharpens its reasoning (via RL) while remaining anchored to the broad support of expert knowledge (via SFT).

\section{Implementation Details}
\label{sec:appendix_implementation}

\subsection{Training Configuration}
Our experiments are conducted using the \textbf{VERL} framework on a cluster of 64 $\times$ NVIDIA H20 GPUs. The training utilizes the Stepwise-GRPO algorithm. Detailed hyperparameters are listed in Table \ref{tab:hyperparams}.

\begin{table}[H] 
\centering
\caption{Hyperparameter settings for the Qwen3-32B Stepwise-GRPO training.}
\label{tab:hyperparams}
\footnotesize 
\renewcommand{\arraystretch}{1.1} 
\begin{tabular}{ll}
\toprule
\textbf{Configuration} & \textbf{Value} \\
\midrule
\multicolumn{2}{l}{\textit{Model \& Optimizer}} \\
Base Model & Qwen3-32B-SFT \\
Precision & bfloat16 \\
Optimizer & AdamW \\
Actor Learning Rate & $1 \times 10^{-6}$ \\
Global Train Batch Size & 256 \\
PPO Mini Batch Size & 256 \\
Max Sequence Length & 5120 \\
Max Prompt Length & 4096 \\
Max Response Length & 1024 \\
\midrule
\multicolumn{2}{l}{\textit{GRPO Parameters}} \\
Number of Rollouts ($N$) & 8 \\
Temperature & 1.0 \\
KL Penalty Coefficient ($\beta$) & 0.001 \\
Advantage Estimator & GRPO \\
Clip Ratio & 0.2 \\
\midrule
\multicolumn{2}{l}{\textit{AFRL Structure Reward Weights}} \\
$w_{\text{decision}}$ (First Token) & 10.0 \\
$w_{\text{trace}}$ (Boxed Steps 4--8) & 10.0 \\
$w_{\text{final}}$ (Final Answer) & 5.0 \\
Format Error Penalty & -1.0 \\
\bottomrule
\end{tabular}
\end{table}
\subsection{Curriculum Learning Schedule}

\begin{table}[H] 
\centering
\caption{Dynamic schedule for Mode-Balanced Optimization coefficients.}
\label{tab:curriculum}
\small 
\renewcommand{\arraystretch}{1.2}
\begin{tabular}{lccc}
\toprule
\textbf{Stage} & \textbf{Hard Ratio} & \textbf{$\alpha_t$ (RL)} & \textbf{$\gamma_t$ (SFT)} \\
\midrule
Stage 1 & 40.0\% & 0.85 & 0.15 \\
Stage 2 & 55.7\% & 0.90 & 0.10 \\
Stage 3 & 65.6\% & 0.95 & 0.05 \\
\bottomrule
\end{tabular}
\end{table}

\end{document}